# An AI-native experimental laboratory for autonomous biomolecular engineering

Mingyu Wu[1†], Zhaoguo Wang[1†]*, Jiabin Wang[2†], Zhiyuan Dong[1], Jingkai Yang[1], Qingting Li[2], Tianyu Huang[1], Lei Zhao[2], Mingqiang Li[2], Fei Wang[2]*, Chunhai Fan[2]*, Haibo Chen[1]*

1. Institute of Parallel and Distributed Systems, Shanghai Jiao Tong University, Shanghai 200240, China

2. School of Chemistry and Chemical Engineering, New Cornerstone Science Laboratory, Frontiers Science Center for Transformative Molecules, National Center for Translational Medicine, Shanghai Jiao Tong University, Shanghai, China.

†These authors contributed equally to this work.

*Correspondence: zhaoguowang@sjtu.edu.cn; wangfeu@sjtu.edu.cn; fanchunhai@sjtu.edu.cn; haibochen@sjtu.edu.cn

**Abstract:**

**Autonomous scientific research, capable of independently conducting complex experiments and serving non-specialists, represents a long-held aspiration. Achieving it requires a fundamental paradigm shift driven by artificial intelligence (AI). While autonomous experimental systems are emerging[1–4], they remain confined to areas featuring singular objectives and well-defined, simple experimental workflows, such as chemical synthesis and catalysis. We present an AI-native autonomous laboratory, targeting highly complex scientific experiments for applications like autonomous biomolecular engineering. This system autonomously manages instrumentation, formulates experiment-specific procedures and optimization heuristics, and concurrently serves multiple user requests. Founded on a co-design philosophy of models, experiments, and instruments, the platform supports the co-evolution of AI models and the automation system. This establishes an end-to-end, multi-user autonomous laboratory that handles complex, multi-objective experiments across diverse instrumentation. Our autonomous laboratory supports fundamental nucleic acid functions—including synthesis, transcription, amplification, and sequencing. It also enables applications in fields such as disease diagnostics, drug development, and information storage. Without human intervention, it autonomously optimizes experimental performance to match state-of-the-art results achieved by human scientists[5]. In multi-user scenarios, the platform significantly improves instrument utilization and experimental efficiency. This platform paves the way for advanced biomaterials research to overcome dependencies on experts and resource barriers, establishing a blueprint for science-as-a-service at scale.**

## Introduction

Developing autonomous "robot scientists" capable of independent discovery and providing end-to-end experimental services to non-experts remains an enduring scientific ambition. Recent AI breakthroughs are accelerating progress toward this goal. Emerging platforms[1,3,4,6–11] integrate AI models like large language models (LLMs)[12–15] with automated hardware, enabling experiment execution through natural language instructions. These systems typically rely on predefined workflows or heuristics from human scientists. For instance, the autonomous platform proposed by Dai et al.[3] can perform simple organic chemistry experiments with predefined procedures, largely fulfilling the role of an organic chemist.

A critical challenge for current autonomous platforms is their inability to understand and perceive the experimental process. They require scientists to manually configure instruments, formulate procedures, and design heuristics for optimization[16–21]. This stems from their "add-on" architecture, where machine learning models merely supplement existing automated hardware rather than integrating natively. Consequently, such systems are confined to tasks with singular objectives and well-defined workflows, like chemical synthesis and catalysis. Their application in scenarios requiring concurrent multi-objective optimization and multi-task collaboration remains severely hindered. Molecular biology laboratories, typified by nucleic acid experimentation[5,22,23], exemplifies this challenge. These experiments require the coordinated operation of diverse, high-precision instruments and iterative optimization of procedures against multiple objectives (e.g., time, yield, and cost). These requirements create vast decision spaces that existing systems cannot navigate. This complexity intensifies in multi-user environments, surpassing current autonomous capabilities.

We introduce an AI-native autonomous laboratory to address these challenges. Our approach replaces conventional add-on architectures with a model-experiment-instrument co-design. This integration enables AI models to precisely comprehend experimental requirements and instrument capabilities, supporting autonomous optimization and decision-making for complex tasks without predefined heuristics. The approach establishes a fully autonomous closed-loop system implementing continuous "design-experiment-optimize" cycles. Furthermore, the laboratory dynamically manages instrumentation and adapts procedures based on real-time instrument status. This capability maximizes both experimental efficiency and instrument utilization when serving multiple users.

Based on this approach, we developed AutoDNA—an AI-native laboratory for nucleic acid science. The system encompasses fundamental functionalities including synthesis, amplification, transcription, and sequencing. It further provides applied services spanning disease diagnostics, drug development, and data storage, accessible through natural language requests from non-experts. For its most complex application, DNA data storage, AutoDNA autonomously coordinates over 20 instruments across more than 9,300 hardware steps. The system can also formulate experimental procedures matching state-of-the-art results achieved by human scientists[5]. In multi-user shared-resource scenarios, AutoDNA's real-time status-aware scheduling enhances experimental throughput by 3 folds compared to conventional approaches.

**AutoDNA system architecture**

AutoDNA features two modules: chemical planning and hardware execution (Fig. 1a, b). Each module incorporates multiple LLM-powered agents[24,25] that communicate through natural language. These agents leverage LLMs' language understanding[26] and reasoning capabilities[27-29] to enable inter-module collaboration and co-evolution. With AutoDNA, a user can specify experimental objectives and requirements through natural language (e.g., 'perform RPA-based sample test'). The chemical planning module then formulates an experimental procedure that actuates the hardware execution module to execute the experiment autonomously. Agents continuously monitors real-time experimental feedback and instrument status, enabling autonomous procedure optimization to enhance experimental yield and efficiency.

The chemical planning module incorporates four specialized agents: 1) Experiment Planner Agent (EPA): Designs experimental procedures based on user requirements; 2) Hypothesis Proposer Agent (HPA): Generates optimization hypotheses for existing procedures; 3) Literature Researcher Agent (LRA): Retrieves key information from scientific literature[30,31] to inform procedure design through EPA collaboration; 4) Reagent Manager Agent (RMA): Manages reagent inventory and enforces resource constraints during procedure generation.

The hardware execution module embodies the "AI-native" philosophy through two specialized agents: 1) Program Developer Agent (PDA): Abstracts instrument functionalities into AI-native interfaces and translates high-level procedures into executable code for hardware operation; 2) Hardware Executor & Validator Agent (HEVA): Deploys code, executes experiments, validates results, and provides feedback to the chemical planning module. On multi-user requests, PDA and HEVA collaboratively manage hardware resources, enabling time-sharing to maximize utilization and throughput.

AutoDNA executes closed-loop experimentation through multi-agent collaboration (Fig. 1a-b). When receiving a user request, the Experiment Planner Agent (EPA) first generates candidate procedures with assistance from the Literature Researcher Agent (LRA). EPA then cross-references the Reagent

Manager Agent (RMA) for available reagents, screening out non-viable procedures. Subsequently, EPA sends the procedures to the hardware execution module, where the Program Developer Agent (PDA) generates corresponding control code. The Hardware Executor & Validator Agent (HEVA) then schedules and runs the code according to real-time instrument status. After execution, HEVA reports the validated results back to EPA, which determines if further optimization is needed. If needed, EPA leverages the Hypothesis Proposer Agent (HPA) to identify improvement pathways, iteratively generating refined procedures until meeting preset objectives. The entire workflow operates through autonomous agent coordination without human intervention.

AutoDNA employs a model-instrument co-design methodology that enables end-to-end autonomy, eliminating manual instrument modeling or control code development. Specifically, the Program Developer Agent (PDA) transforms instrument functionalities—extracted from technical documentations—into AI-native abstractions called 'atomic services'. For example, a thermal cycler is abstracted as two atomic services: temperature regulation (set_temp) and reaction initiation (start) services. Each atomic service is represented as a Python object containing natural language descriptions, aligning with agents' natural language understanding capabilities. PDA further autonomously validates code integrity and performs error correction using instrument documentation.

Unlike existing approaches that depend on predefined heuristics, AutoDNA enables autonomous multi-objective optimization through multi-agent collaboration. The Experiment Planner Agent (EPA) orchestrates this process with the Literature Researcher (LRA) and Hypothesis Proposer Agent (HPA). LRA identifies optimal procedures through literature analysis. When experimental results underperform, HPA analyzes data to generate improvement hypotheses. EPA then integrates these hypotheses with LRA's literature insights to formulate revised procedures. For multi-objective tasks, EPA prioritizes objectives according to user specifications. For example, in an enzymatic DNA synthesis experiment with suboptimal yield, HPA proposes 'suboptimal enzymatic reaction efficiency' as a causal hypothesis. LRA then identifies potential factors (e.g., reaction temperature or reagent concentration) from literature, enabling EPA to design an optimal procedure.

AutoDNA also features autonomous hardware management capabilities, enabling the concurrent processing of multiple experimental requests. This significantly improves instrument utilization and experimental efficiency compared to existing autonomous platforms. The Hardware Executor & Validator Agent (HEVA) maintains real-time awareness of instrument status and generates optimal resource scheduling strategies. This design decouples control code generation from physical resource scheduling: the Program Developer Agent (PDA) generates control code assuming exclusive instrument access, while HEVA consolidates and schedules requests from multiple users. On instrument conflicts, HEVA and PDA collaboratively reroute procedures to functionally equivalent instruments. For example, when a DNA library preparation experiment requires heating but the heater is occupied, PDA dynamically adapts code to perform isothermal incubation on an available thermal cycler, enabling efficient parallel experimentation.

We implemented AutoDNA on a physical platform for general-purpose nucleic acid experimentation (Fig. 1c). The system connects diverse instruments through pipettes and robotic arms, enabling reagent transfer between any two instruments. The hardware achieves microliter-scale volumetric precision and micrometer-scale positional accuracy—meeting molecular biology's low-volume, high-precision requirements. The platform supports a comprehensive range of nucleic acid workflows including synthesis, amplification, transcription, and sequencing. This versatility enables end-to-end services for disease diagnostics, information storage, and drug development applications.

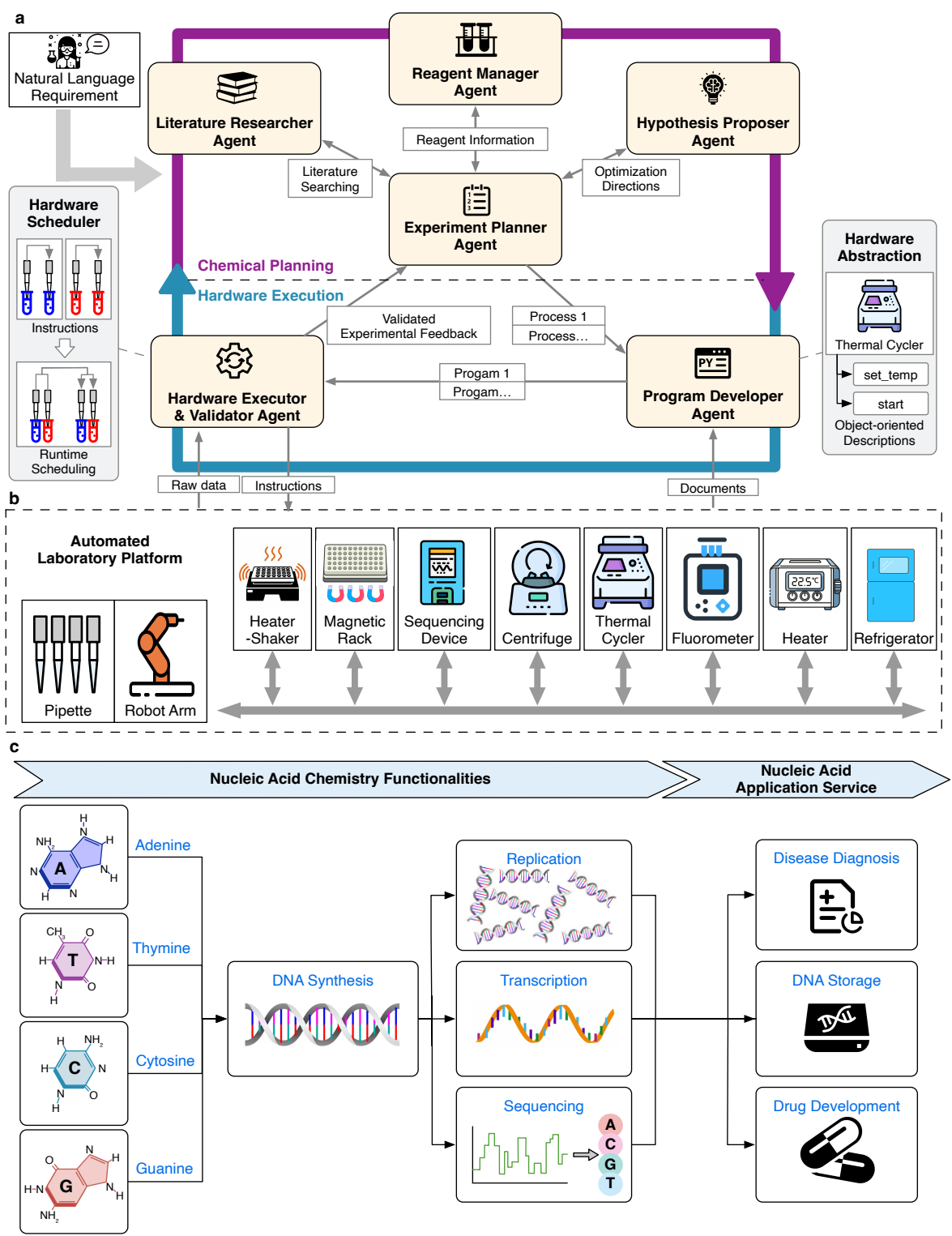

**Fig 1. The architecture of AutoDNA, an AI-native autonomous laboratory. a,** The platform with a multi-agent architecture where specialized agents collaboratively execute closed-loop "design-experiment-optimize" workflows. **b,** The automated hardware platform, which integrates instruments common to nucleic acid manipulation and co-designs with the agents. **c,** Representative fundamental nucleic acid functionalities and application services supported by AutoDNA.

**End-to-end autonomous nucleic acid test service**

We first demonstrated AutoDNA's end-to-end autonomous service capabilities through a nucleic acid test application, enabled by model-instrument co-design. Nucleic acid test identifies specific genetic sequences (RNA/DNA) and is pivotal for viral diagnostics, genetic disease screening, and environmental monitoring. A common nucleic test method is Recombinase Polymerase Amplification (RPA)[23]. It utilizes recombinase and polymerase to achieve rapid, isothermal amplification, and is characterized by its fast kinetics and procedural simplicity (Fig. 2a).

As a typical workflow for this task, the Experiment Planner Agent (EPA) directly generated a three-step workflow given RPA's procedural simplicity—comprising reagent mixing, incubation, and fluorescence test—without reliance on literature retrieval (Fig. 2b). We then evaluated code generation with versus without hardware abstractions. Without abstractions, the Program Developer Agent (PDA) produced only abstract pseudocode devoid of instrument-specific commands. Conversely, when leveraging hardware abstractions, PDA generated executable instrument code coordinating robotic arms, pipettes (reagent transfer), thermal cyclers (temperature control), and fluorometers (test). This enabled autonomous experiment execution.

Notably, the Program Developer Agent (PDA) autonomously inferred the need to seal reaction tubes during incubation, invoking the robotic arm's lid closing operation (capTubeGroup). This indicates that the agent's underlying LLM recognized the implicit requirement for a closed environment in RPA (preventing evaporation) and could reason with the hardware abstractions to enhance experimental fidelity. Furthermore, PDA detected and corrected a critical code flaw guided by the instrument operation descriptions: a missing reagent transfer step that would have activated thermal cyclers and fluorometers on empty wells.

Fluorescence measurements from autonomous experiments showed statistically equivalent values to manual controls for identical samples (Fig. 2c). Critically, both methods achieved diagnostic concordance in sample positivity determination, confirming AutoDNA's ability to deliver end-to-end nucleic acid test from simple high-level natural language instructions.

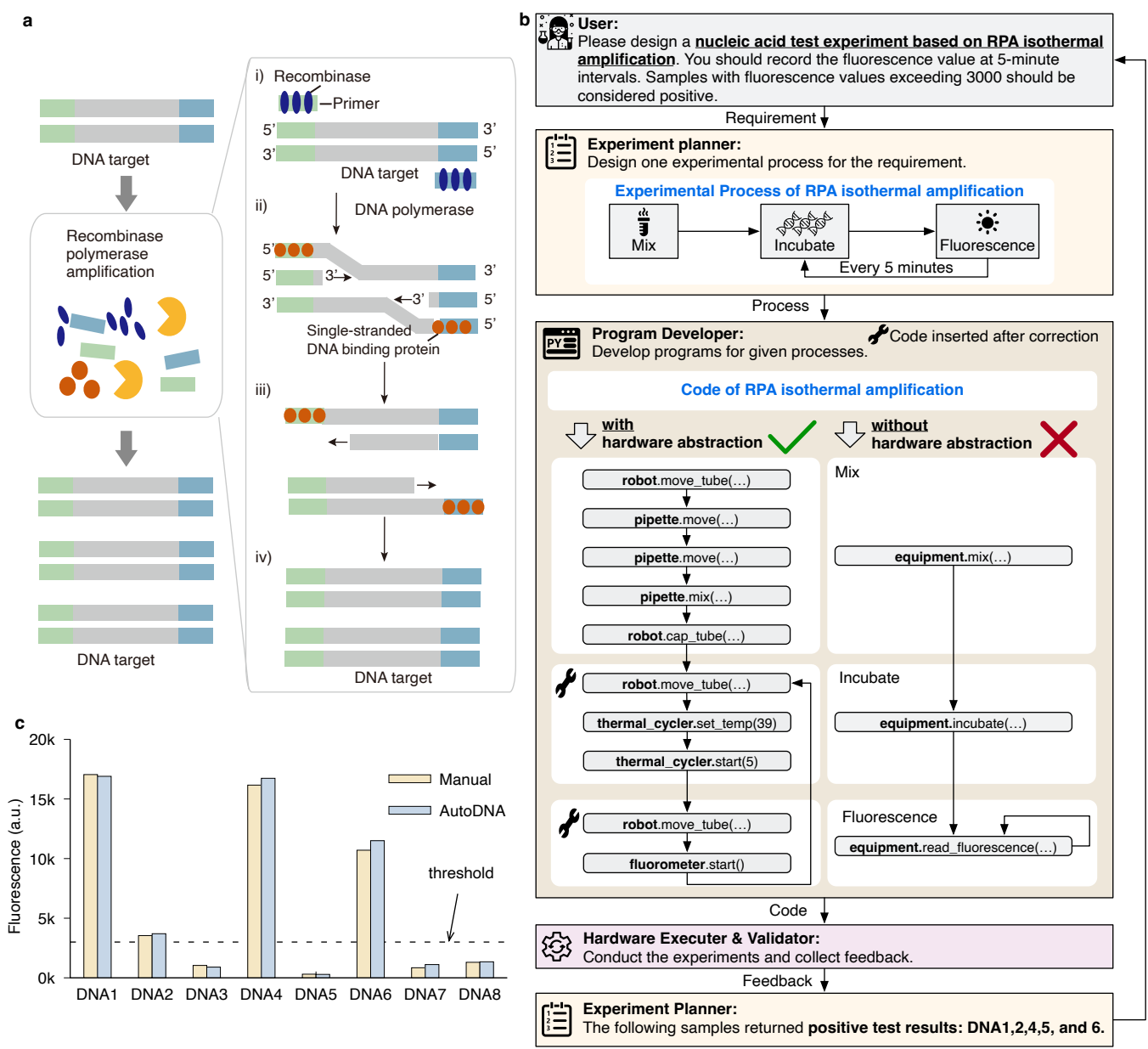

**Fig 2. Hardware abstractions enable autonomous code generation for a nucleic acid test service by AutoDNA. a,** Schematic illustration of a common process of RPA-based nucleic acid test. **b,** The autonomous workflow for the RPA-based nucleic acid test service, contrasting the code generation outcomes with and without hardware abstractions. **c,** Comparison of nucleic acid test results from the agent-driven autonomous experiment and a corresponding manual one.

## Multi-objective autonomous optimization services for enzymatic DNA synthesis

Next, we demonstrated AutoDNA's autonomous multi-objective optimization using enzymatic de novo DNA synthesis. Compared to conventional phosphoramidite synthesis[32], enzymatic DNA synthesis offer superior coupling efficiency, milder reaction conditions, and the ability to produce longer DNA strands, positioning it as the next-generation DNA synthesis technology[33,34]. The core process alternates between two steps (Fig. 3a): 1) Single-nucleotide extension catalyzed by an engineered terminal deoxynucleotidyl transferase (TdT) enzyme[5], adding reversible nucleotide terminators to growing DNA strands, and 2) Deblocking to remove protecting groups for strand elongation. While this iterative cycle enables custom sequence synthesis, non-standardized reagents and reaction conditions create a large and complex search space for optimization.

AutoDNA generates an iterative workflow to serve a synthesis request optimizing for maximum yield and minimal execution time. The Experiment Planner Agent (EPA), assisted by Literature Researcher Agent (LRA), first identified two candidate buffers for the post-reaction washing step: lysis buffer and B&W buffer. Based on experimental feedback from the hardware execution module, EPA selected the B&W buffer for subsequent experiments (Fig. 3c). As the synthesis yield was still suboptimal, EPA engaged the Hypothesis Proposer Agent (HPA) implementing further optimization strategies: the agents decided to introduce the surfactant Tween-20 and increase the concentrations of the $CoCl_2$ cofactor, the TdT enzyme, and the reversible terminators. These adjustments successfully increased the yield to over 98%. EPA then shifted its focus to minimizing the reaction time, but this led to a decrease in yield, prompting the agent to halt further optimization. In summary, AutoDNA explored optimization across diverse dimensions—including buffers, surfactants, reaction time, and reagent concentrations—iteratively validating its decisions and collecting feedback from experiments on the hardware platform. In contrast to existing methods that require predefined optimization spaces[1,3,35] or fixed experimental workflows[2,4,6,7,36], AutoDNA's optimization search space is autonomously explored and constructed by agents. This approach not only eliminates the need for manual intervention but also enables the discovery of optimization dimensions that human experts might overlook.

Polyacrylamide gel electrophoresis (PAGE) analysis for a specified sequence (3'-GTGTGTGT-5') after autonomous optimization iteratively performed by the agents revealed a step-wise average yield of 97.7% (Fig. 3d), which is comparable to previously reported manually optimized results[5]. We further characterized the error profile of the synthesized product using Nanopore sequencing (Fig. 3e). The primary error source was deletion, accounting for 2.35% of total, which we hypothesize is mainly caused by magnetic bead aggregation. The error rates for insertion and substitution were 0.25% and 0.12%, respectively. These results demonstrate AutoDNA's ability to autonomously navigate a complex search space and achieve expert-level performance in a multi-objective synthesis task.

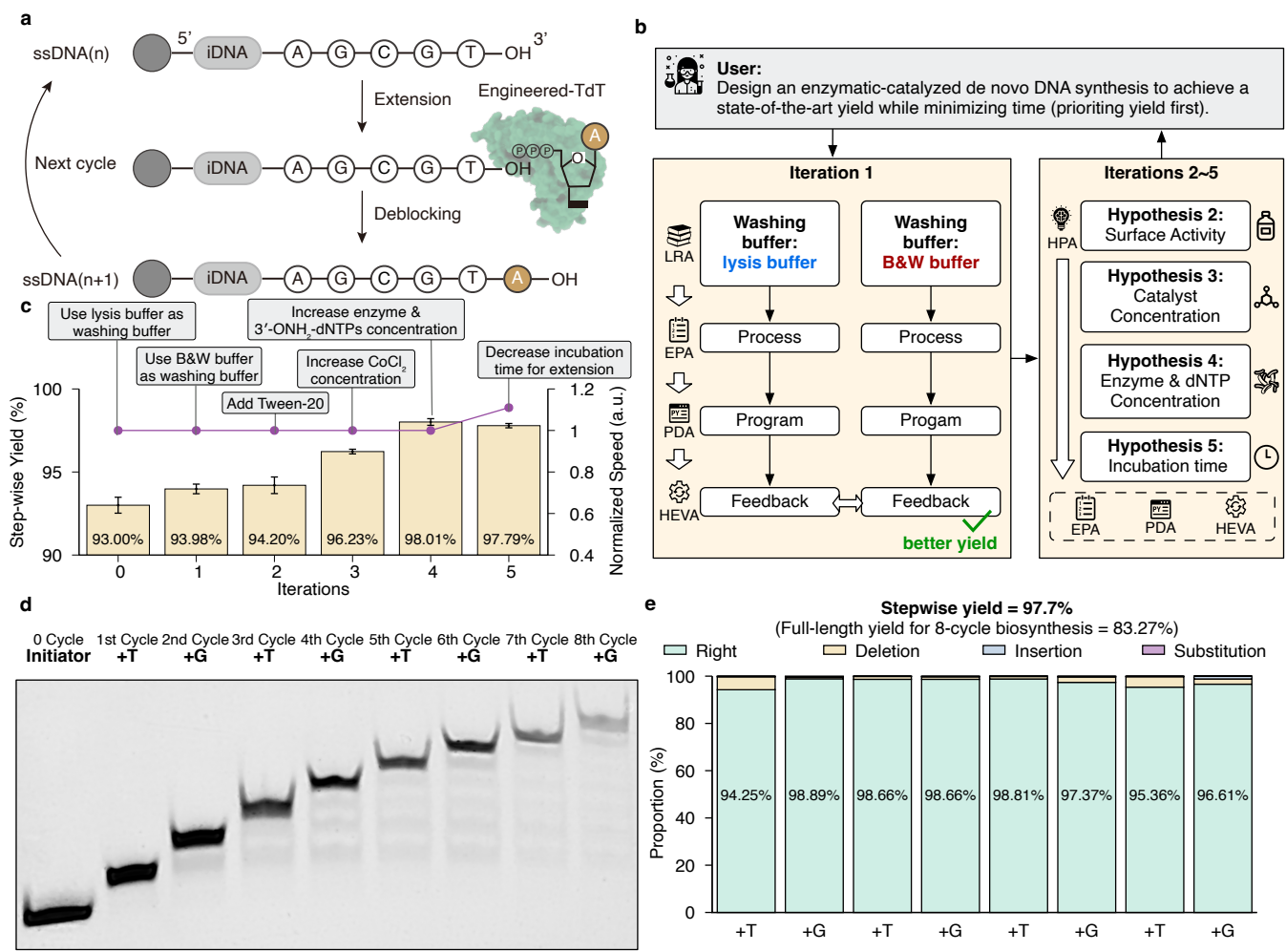

**Fig 3. Autonomous workflow for the optimization of enzymatic DNA synthesis by AutoDNA. a,** Schematic of the enzymatic DNA de novo synthesis cycle. **b,** Workflow of agents' autonomous searching and optimization process. **c,** Trajectory of the multi-objective optimization (yield and time) across iterative cycles, highlighting key decisions made by agents. **d,** Per-cycle PAGE analysis of the synthesized 3'-GTGTGTGT-5' oligonucleotide. **e,** Estimates of the stepwise yield and error proportion.

**Concurrent multi-request support**

We further evaluated the capacity of AutoDNA for autonomous instrument management across two multi-experiment scenarios. The first scenario addresses parallel requests from a single user, where agents may generate multiple viable procedures. For example, during the enzymatic synthesis optimization, the Experiment Planner Agent (EPA) generated two similar, feasible procedures in the first iteration. The Hardware Executor & Validator Agent (HEVA) then merged these procedures for concurrent execution. Taking the pipetting-related code as an illustration (Fig. 4a), the Program Developer Agent (PDA) only specifies which reagents to transfer. During consolidation, HEVA maps these abstract commands to a physical layout—a horizontal array of tubes—enabling the multi-channel pipetting to perform transfers efficiently in a unified motion. This parallelization strategy is extended to subsequent steps such as deprotection, addition, and washing. This optimized scheduling by HEVA reduced the overall exploration time by 3.60X compared to sequential execution (from 1,434.7 minutes to 398.7 minutes; Fig. 4d).

The second scenario simulates a dynamic, multi-user environment where three users concurrently submit requests for DNA poly(A) tailing, DNA library preparation, and nucleic acid test. Initially, the Program Developer Agent (PDA) scripted an isothermal reaction for the library preparation using a heater, but HEVA identified that this instrument was already allocated to the poly(A) tailing experiment. After receiving the feedback, PDA leveraged the hardware abstraction to recognize the functional equivalence of a thermocycler, re-scripting the procedure to use the available instrument instead (Fig. 4b). The resulting instrument utilization trace (Fig. 4c) shows that the dynamic scheduling improves the efficiency of both the heater and the thermocycler (Fig. 4f) and achieves a time saving of 168.8 minutes compared to a serial, queue-based execution (Fig. 4e).

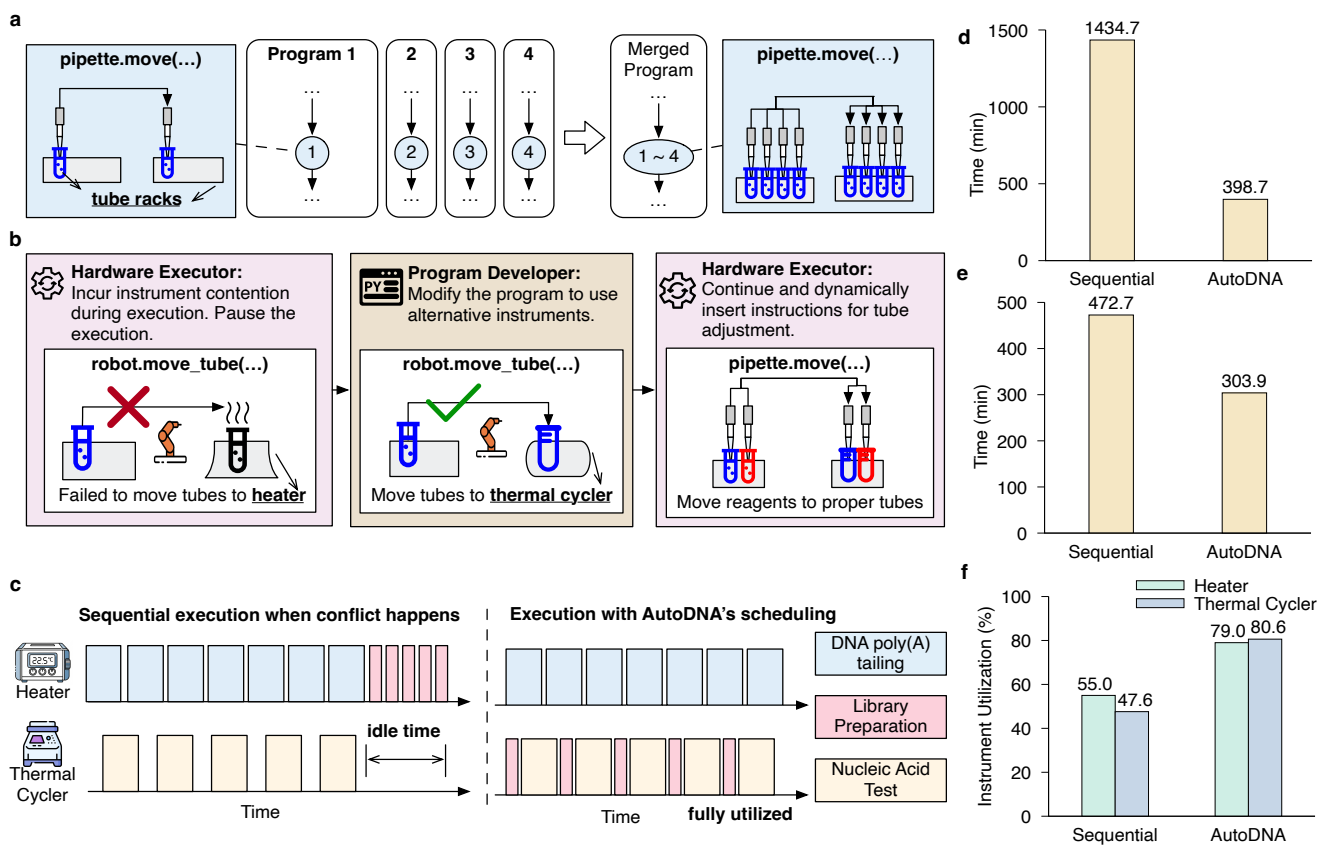

**Fig 4. AutoDNA's management of concurrent experiment requests. a,** Consolidation and scheduling of multiple control programs by HEVA for concurrent execution, illustrated with a pipetting operation. **b,** Collaborative code refinement by PDA and HEVA to resolve an instrument conflict in a multi-user scenario. **c,** Instrument utilization trace for the heater and thermocycler, corresponding to the conflict resolution shown in **b**. **d,** Reduction in experiment time for the enzymatic synthesis service achieved by optimized scheduling. **e,** Overall time savings in the multi-user scenario achieved through dynamic conflict resolution and scheduling. **f,** Improvement on the utilization rate of two involved instruments.

## Service integration for DNA storage

AutoDNA further handles complex workflows by autonomously decomposing them into executable subtasks. We demonstrated this capability using an end-to-end DNA data storage service[37,38]. As the carrier of genetic information, DNA is a promising next-generation medium for large-scale data storage owing to its high information density, long-term stability, and low energy consumption for preservation[22,39,40]. A typical in vitro DNA storage workflow involves five key steps (Fig. 5a): 1) Encoding, converting digital information into nucleotide sequences; 2) Writing, the de novo synthesis of the specified DNA strands; 3) Storing, preserving the synthesized DNA; 4) Reading, retrieving the sequence information via sequencing; and 5) Decoding, translating the sequenced data back to its original format. This workflow also requires intermediate steps, such as library preparation before sequencing.

AutoDNA autonomously executed the entire workflow (Fig. 5b). From a high-level user command specifying data and operation (read or write), the Experiment Planner Agent (EPA) autonomously decomposes the request into subtasks. The encoding and decoding are performed in silico, which EPA completes using its existing knowledge. For the wet-lab synthesis and sequencing steps, the agents apply the methodology described previously: performing literature retrieval, reagent queries, procedure generation, code generation, and hardware execution. As the enzymatic de novo synthesis procedure had been generated in previous experiments, we archived it in the system's knowledge base, allowing agents to retrieve and execute it directly for the writing step.

AutoDNA complete the full read-write cycle in 162.9 hours, orchestrating 25 instruments across 9,365 hardware steps (where each invocation of an atomic service counts as one step), all without any manual intervention except for reagent replenishment. All 78 synthesized strands were read with high accuracy, and the information was automatically recovered and decoded to the same content written (Fig. 5b). Similar to previous enzymatic synthesis experiments, the primary error remained deletions (Fig. 5c), which we hope to address in the future with further improvements to the magnetic bead aggregation issue.

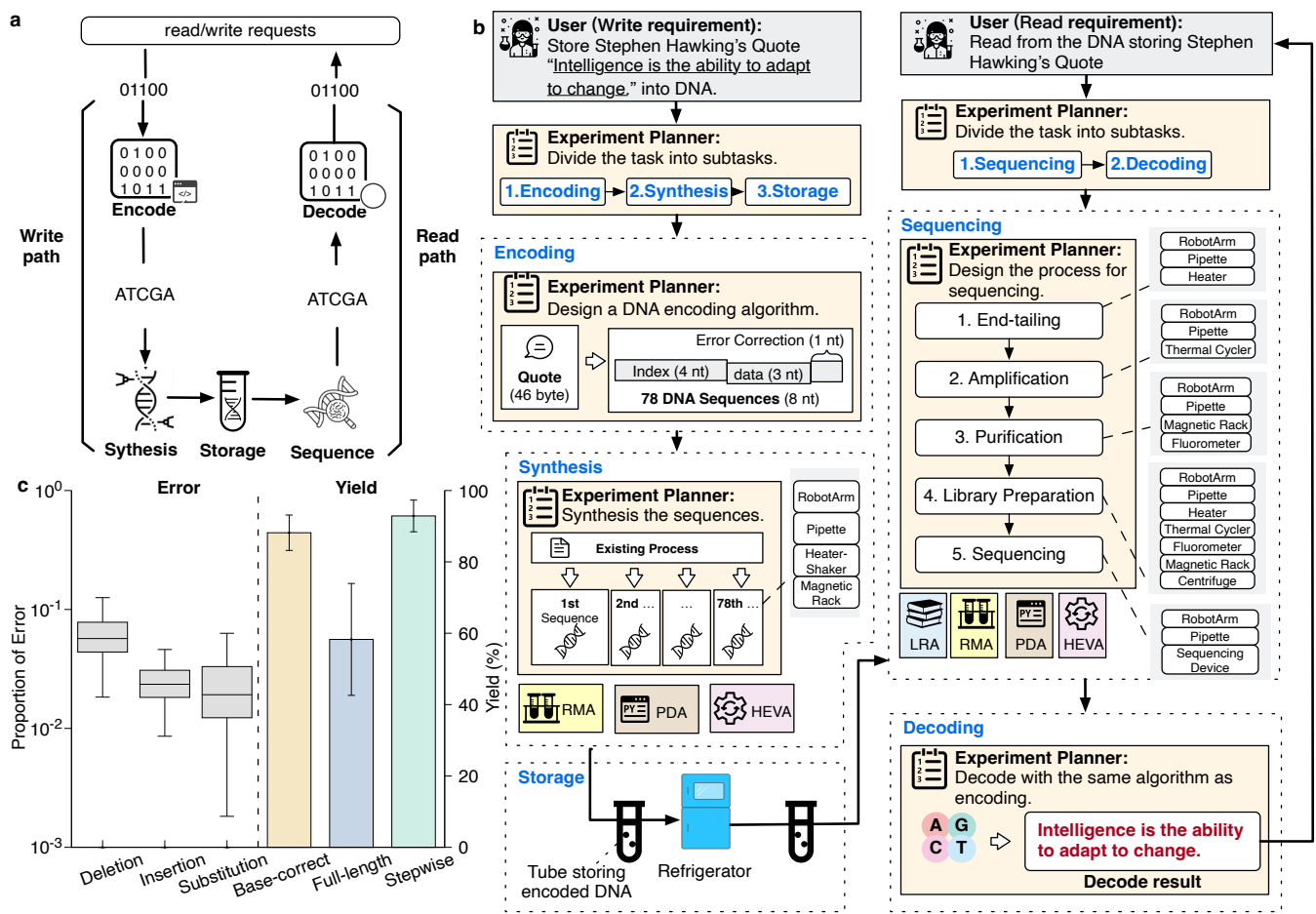

**Fig 5. An end-to-end DNA data storage service supported by AutoDNA**. **a**, Schematic of the end-to-end DNA data storage workflow. **b**, The agents' process for autonomously decomposing a high-level natural language request into an executable workflow for DNA writing and reading. **c**, Analysis results of the average yield and error rate, according to Nanopore sequencing.

## Discussion

We introduced an AI-native autonomous experiment laboratory capable of executing complex scientific experiments (e.g., nucleic acid experiments) without the need of human intervention. The system delivered end-to-end services for non-experts through an easy-to-use natural language interface. By seamlessly coordinating over 20 instruments and autonomously optimizing procedures, it achieved experimental yields matching expert optimized procedures while tripling parallel throughput versus conventional systems. This capability supported integrated services executing more than 9,000 discrete hardware steps.

In contrast to existing approaches[1-4,35] requiring human chemists to define experimental procedures or decision heuristics, our AI-native approach—centered on the synergistic co-design of the model, experiment, and instruments—is key to achieving fully autonomous experimentation. Three innovations enable such full autonomy. First, the agents provide intelligent extensions to the hardware platform (the Program Developer and Hardware Executor & Validator Agent), creating a robust interface between the physical world of hardware and the digital world of AI models. This establishes the foundation for agents to understand instrument operation and automate experiments. Second, the multi-agent architecture adheres to the principles of the scientific method, leveraging the reasoning capabilities of LLMs to extract knowledge from literature and formulate experimental hypotheses. This obviates the need for pre-configured heuristics, opens up richer dimensions for optimization, enabling a closed-loop "design-experiment-optimize" cycle that rivals the performance of manual optimization. Third, when handling concurrent experiment requests, the agents dynamically monitor hardware status to orchestrate concurrent experiments, thereby enhancing instrument utilization and overall efficiency.

Our future work will further explore co-design methodologies for AI and automation platforms. We aim to extend the scope of autonomous experimentation to more exploratory domains like advanced biomaterials discovery, with the ultimate goal of enabling AI-driven scientific discovery. Concurrently, we will scale platform capabilities to serve broader non-expert communities.